\pdfoutput=1

\documentclass[11pt]{article}

\usepackage[]{emnlp2021}

\usepackage{times}
\usepackage{latexsym}

\usepackage[T1]{fontenc}

\usepackage[utf8]{inputenc}

\usepackage{microtype}

\usepackage{array}
\usepackage{graphicx}
\usepackage{clrscode}
\usepackage{balance}
\usepackage{subfigure}
\usepackage{multirow}
\usepackage{booktabs}
\usepackage{adjustbox}
\usepackage[justification=justified,skip=5pt]{caption}
\usepackage{float}
\usepackage{color}
\usepackage{soul}
\usepackage{mdframed}
\usepackage{amsopn}
\usepackage{mathrsfs}
\usepackage{mathtools}
\usepackage{amsmath}
\usepackage{arydshln}
\usepackage{hyperref}
\usepackage{multicol}
\usepackage{blkarray}
\usepackage{enumerate}
\usepackage{courier}
\usepackage{rotating}
\usepackage{booktabs}
\usepackage{diagbox}
\usepackage{fancybox}
\usepackage{minibox}
\usepackage{cases}
\usepackage{bm}
\usepackage[lined, ruled, commentsnumbered, linesnumbered]{algorithm2e} %
\usepackage{enumitem}
\title{Counterfactual Evaluation for Explainable AI}

\author{Yingqiang Ge, Shuchang Liu, Zelong Li, Shuyuan Xu, Shijie Geng,\\ 
    {\bf  Yunqi Li, Juntao Tan, Fei Sun$^{\dagger}$, Yongfeng Zhang} \\
    \texttt{Rutgers, The State University of New Jersey} \\
    $^{\dagger}$ \texttt{Alibaba Group}\\
    \text{\{yingqiang.ge,sl1471,zelong.li,shuyuan.xu,shijie.geng\}@rutgers.edu}\\
    \text{\{yunqi.li,juntao.tan\}@rutgers.edu},ofey.sunfei@gmail.com, yongfeng.zhang@rutgers.edu\\

  }

\begin{document}
\maketitle
\begin{abstract}
While recent years have witnessed the emergence of various explainable methods in machine learning, to what degree the explanations really represent the reasoning process behind the model prediction---namely, the faithfulness of explanation---is still an open problem.
One commonly used way to measure faithfulness is \textit{erasure-based} criteria.
Though conceptually simple, erasure-based criterion could inevitably introduce biases and artifacts.
We propose a new methodology to evaluate the faithfulness of explanations from the \textit{counterfactual reasoning} perspective: the model should produce substantially different outputs for the original input and its corresponding counterfactual edited on a faithful feature.
Specially, we introduce two algorithms to find the proper counterfactuals in both discrete and continuous scenarios and then use the acquired counterfactuals to measure faithfulness.
Empirical results on several datasets show that compared with existing metrics, our proposed counterfactual evaluation method can achieve top correlation with the ground truth under different accessing conditions to the black-box model's outputs.

\end{abstract}

\section{Introduction}\label{sec: intro}


Despite the emergence of algorithmic approaches to explaining
machine learning models~\cite{doshi2017towards,hase2020evaluating,jacovi2020towards}, one critical issue is the challenge of defining and evaluating what constitutes a quality explanation.
\citet{jacovi2020towards} distinguished between distinct aspects of the explanation's quality, especially, the \textit{plausibility} and \textit{faithfulness}.
Measuring the agreement between human-generated and algorithm-generated explanations assesses the plausibility of explanations (agreeable to humans), while faithfulness focuses on whether the model actually relied on these particular explanations to make a prediction (agreeable to the model)~\cite{deyoung2019eraser,jacovi2020towards}. 
While human judgment can be used to assess plausibility, it may not be used to evaluate faithfulness since if humans understood the model, explaining the models would be pointless, which makes faithfulness assessment even more complicated and difficult.



In literature, a widely employed criterion to measure faithfulness is \textit{erasure-based} criteria, with the key insight that erasing the most (or least) important features should lead to the most (or least) significant function value difference\footnote{This metric is often employed on methods that consider heat-maps (e.g., attention maps) over the input as explanations, also referred to as feature-attribution explanations~\cite{kim2018interpretability}. So is this work.}~\cite{zaidan2007using,deyoung2019eraser,yu2019rethinking}.
However, erasure-based criteria would cause an intrinsic bias for evaluation as the deletion nature of erasure-based metrics will inevitably introduce ambiguities since multiple different inputs could map to an exactly same input value/structure after erasure, making the evaluation fuzzy.
For example, for a sentence $A=\{\dots\text{context}\dots, \text{``good''}\}$ that is originally classified as positive and a sentence $B=\{\dots\text{context}\dots, \text{``bad''}\}$ that is classified as negative and only differs from $A$ on the last word.
Suppose that the last word of both sentences are selected as explanations.
After erasure, both sentences will reduce to the same sentence with only the context.
If the context is classified as positive/negative, then the erasure-based evaluation will determine that the explanation model fails on sentence A/B.
Besides, in cases where faithful explanations are required, an inaccurate or misleading evaluation can have catastrophic consequences~\cite{jacovi2020towards}.
To overcome these problems, in this work, we propose a new evaluation method from the perspective of \textit{counterfactual reasoning}.

The counterfactual evaluation process is carefully designed based on the following two intuitive observations:
\begin{itemize}
    \item Even small perturbations on the most important features may influence the model prediction;
    \item Unless the perturbations on the least important features are large enough, they will not influence the model prediction too much.
\end{itemize}

Based on these two observations, we propose a new framework leveraging the notion of counterfactual to evaluate the faithfulness of explanations.
In the counterfactual evaluation process, we aim to ask: how the model's prediction would change, had we changed its input on identified features.
Specially, we first propose different algorithms to find the proper counterfactuals when dealing with discrete or continuous input conditions, respectively.
Then, we use \textit{Proximity}---the average edit distance between original inputs and the corresponding counterfactuals---to measure the degree of perturbations, and use \textit{Validity}---the ratio of the counterfactuals that actually have the desired outputs (e.g., label changed in classification task) over the total number of data points---to measure the degree of prediction switches \cite{moraffah2020causal, verma2020counterfactual,mothilal2020explaining}.
Moreover, inspired by~\citet{deyoung2019eraser}, we also define $\textit{Validity}_{\mathrm{soft}}$---the average change from the original probabilities to the predicted probabilities on the same class with counterfactual inputs---to measure the degree of probability changes.
Finally, based on different accessing conditions to the black-box model's outputs, we combine \textit{Proximity} with \textit{Validity} or with $\textit{Validity}_{\mathrm{soft}}$ to acquire the Counterfactual Evaluation Score $\mathcal{C}$ or $\mathcal{C}_{\mathrm{soft}}$ to measure the faithfulness of explanations.
The advantages of the proposed Counterfactual Evaluation (CE) method can be summarized as follows:
\begin{enumerate}
    \item CE is a generalization of erasure-based criteria based on counterfactual reasoning;
    \item Since CE uses the counterfactual samples, it reduces the bias in erasure-based evaluation measures that focus on ``removing features'';
    \item $\mathcal{C}$ or $\mathcal{C}_{\mathrm{soft}}$ can be used to measure faithfulness under different accessing conditions to the black-box model's outputs.
\end{enumerate}


We apply our counterfactual evaluation method to automatically evaluate and compare a series of explanation approaches, e.g., LIME~\cite{ribeiro2016should}, Anchor~\cite{ribeiro2018anchors}, and so on, through the comparisons with ground-truth explanations provided by a whitebox model, Logistic Regression~\cite{molnar2020interpretable,ribeiro2016should}.
The superiority of our proposed counterfactual evaluation process is supported by empirical results on several datasets, showing that $\mathcal{C}$ or $\mathcal{C}_{\mathrm{soft}}$ can achieve top correlations with the ground-truth when comparing with the established metrics under various accessing conditions to the black-box model's outputs.


\section{Related Work}

\subsection{Explainability in Machine Learning}

Explanation methods can be broadly categorized into two categories---inherently interpretable models and post-hoc methods.
Inherently interpretable models can generate explanations in the process of decision making~\cite{molnar2020interpretable,10.1145/3400051.3400058}, e.g., decision tree, linear regression, logistic regression, and attention networks~\cite{DBLP:journals/corr/VaswaniSPUJGKP17,velickovic2018graph}, etc.
Post-hoc interpretable methods aim to explain the decision-making process of the trained black-box models, and the most prominent solutions use gradient-based approaches~\cite{simonyan2014deep,li-etal-2016-visualizing} which assumes that the black-box is differentiable.
There exist methods that relax this constraint such as Local Interpretable Model-agnostic Explanations (LIME)~\cite{ribeiro2016should}, Anchor explanation (Anchor)~\cite{ribeiro2018anchors} and Structured-Output
Causal Rationalizer explanations (SOCRAT) \cite{alvarezmelis2017causal}.
Their main idea is to observe and approximate the model behavior around the given input.
\citet{goyal2019counterfactual} proposed a post-hoc counterfactual visual explanation method.
Our evaluation framework works on both inherently interpretable models and post-hoc explanation models as long as the model belongs to the feature-attribution explanation family.
Note that there do exist methods that do not employ feature-attribution explanation, for example, prototype/example-based explanation methods use examples (synthetic or natural) to explain individual predictions~\cite{koh2017understanding}.


\subsection{Evaluating Faithfulness}
Having extracted an explanation, researchers also need to evaluate how well the explanation is for a model's prediction.
There exist two particularly notable criteria recently, plausibility and faithfulness.
Plausibility refers to how convincing the interpretation is to humans, while faithfulness refers to how accurately the explanation reflects the true reasoning process of the model~\cite{herman2019promise,wiegreffe-pinter-2019-attention,jacovi2020towards}.
Typically, a plausible explanation may not always be faithful, and a faithful explanation is accurate but may not be easy to understand by human.

While plausibility can be assisted with human study, how to measure faithfulness is still an open question to the community~\cite{deyoung2019eraser}.
Prior studies had compared and analyzed the explainability of feature-attribution methods to identify ``important'' tokens by using erasure as their basis~\cite{nguyen-2018-comparing,serrano-smith-2019-attention,chrysostomou2021variable}.
The intuition is to measure how much the model performance would drop after the set of the ``most important'' tokens/features in an explanation is removed~\cite{zaidan2007using,yu2019rethinking}.
For example, \citet{serrano-smith-2019-attention} measured the percent of decision flip after zeroing the attention weight of a certain element.
Recently, \citet{deyoung2019eraser} proposed two measures of \textit{comprehensiveness} and \textit{sufficiency} as a formal generalization of erasure-based criteria: the degree to which the model is affected by excluding the most (least) important features, or solely including them.
However, the erasure-based criterion has limitations as pointed out by prior work~\cite{bastings-filippova-2020-elephant,ancona2018towards,pruthi2020evaluating}: the corrupted version of an input produced by this criterion might fall out of the data distribution that models are trained on and result in an inaccurate measurement of faithfulness.





One recent work on faithfulness evaluation ~\citet{ding2021evaluating} was based on the model assumption and the prediction assumption proposed by~\citet{jacovi2020towards}, which do not suit our problem formulation.
Another work \citeauthor{ye2021evaluating} also used the idea from counterfactual by acquiring proper counterfactuals by hand.
This limits the evaluation to small scale, and thus unrealistic for large-scale datasets.

\section{\mbox{The Counterfactual Evaluation Method}}

\subsection{Problem Formulation}
Given a black-box model $\mathcal{F}$ (e.g., a classifier) and $K$ feature-based explanation models $\mathcal{E}_1$, $\mathcal{E}_2$, $\dots$, and $\mathcal{E}_K$ (e.g., LIME~\cite{ribeiro2016should}), we want to evaluate whether these explanation models interpret $\mathcal{F}$ faithfully on the evaluation dataset $D=\{\mathbf{x}_1, \mathbf{x}_2, ..., \mathbf{x}_N\}$.
For the remainder of the paper, each data point $\mathbf{x}_i=\{v_{1}=x_{i,1}, \dots, v_{M}=x_{i,M}\}$ is represented as a set of $M$ features $\{v_1,\dots,v_M\}$ with corresponding values $\{x_{i,1},\dots,x_{i,M}\}$.
And for each record $\mathbf{x}_i$ ($i\in\{1,\dots,N\}$), the prediction model $\mathcal{F}$ produces an output $\hat{y}_i = \mathcal{F}(\mathbf{x}_i)$.
In the case of classification, the black-box model first outputs the class probabilities then pick the class with the maximum probability: $\hat{y}_i=\arg\max_{y}p(y|\mathbf{x}_i)$.
Then, for each pair $(\mathbf{x}_{i}, \hat{y}_{i})$, all explanation models provide their explanations $\{\mathbf{e}^i_1, \mathbf{e}^i_2, \dots, \mathbf{e}^i_K\}$ correspondingly.
Note that feature-attribution explanation assumes that features in input $\mathbf{x}_i$ are used to construct the explanation (e.g., a word sequence [``This'', ``is'', ``delightful''] is classified as positive sentiment, and the explanation may provide the word (feature) ``delightful'' as the reason), so $\forall k\in\{1,\dots,K\},\mathbf{e}^i_k \subseteq\mathbf{x}_i$.
In the following, we will first define the standard process of counterfactual evaluation for explainable models in Section \ref{sec:ce}, then specify its discrete and continuous variants in Section \ref{sec:discrete} and Section \ref{sec:continuous},
and discuss the relationship between erasure-based criteria and counterfactual evaluation in Section \ref{sec: cs_vs_erasure}.

\subsection{Counterfactual Evaluation (CE)}\label{sec:ce}

As described in Section \ref{sec: intro}, the goal of the counterfactual evaluation process is to estimate if the explanation method can pick features that can easily switch the output of $\mathcal{F}$.
The whole process consists of two steps: 1) obtain a counterfactual for each input-explanation pair, and then 2) calculate the counterfactual evaluation metrics.

For a given input $\mathbf{x}_i$ and the selected features $\mathbf{e}_k^i$, we denote the \textit{counterfactual} of the original input as $\mathbf{x}_i^{\mathrm{cf}}$, and its corresponding prediction can be inferred by $\hat{y}_{i}^{\mathrm{cf}}=\mathcal{F}(\mathbf{x}_i^{\mathrm{cf}})$.
Now, suppose we have acquired the tuple $(\mathbf{x}_i,\hat{y}_i,\mathbf{e}^i_k,\mathbf{x}_i^{\mathrm{cf}},\hat{y}_i^{\mathrm{cf}})$ for all of the $N$ data points in $D$, the next evaluation step first employ two metrics, Validity and Proximity~\cite{moraffah2020causal, verma2020counterfactual,mothilal2020explaining}, to measure the faithfulness of the corresponding explainable model $\mathcal{E}_k$:\\
\textbf{Validity}: 
this measures the degree of prediction switch induced by the counterfactual on selected features $\mathbf{e}_k^i$. 
It is estimated as the ratio of the number of records that actually changed the class label over the total number of records:
\begin{equation}
\small
\textit{Validity} = 
\frac{1}{N} \sum_{i=1}^{N} I(\hat{y_{i}} \neq \hat{y_{i}}^{\mathrm{\mathrm{cf}}})
\end{equation}
A higher validity is preferable, because it means that more important features are selected in $\mathbf{e}_k^i$.
Recall that a small change on an important feature would likely cause a label change, and if the explanation $\mathbf{e}_k^i$ has almost no effect on the prediction, then no matter how we intervene it, we barely get a prediction switch ($ \hat{y_{i}}^{\mathrm{\mathrm{cf}}} \neq \hat{y_{i}} $).
Note that we are evaluating the explainability of explanation methods rather than the accuracy of the black-box model $\mathcal{F}$, as a result, it does not involve the gold-standard label $y_i$ and only uses the predictions $\hat{y}_i$ and $\hat{y}_i^{\mathrm{cf}}$.\\
\textbf{Proximity}: 
this measures the difficulty of obtaining a counterfactual $\mathbf{x}_i^{\mathrm{\mathrm{cf}}}$ that switches the output.
It is estimated by the average distance between the original input and the counterfactual:
\begin{equation}
\small
\textit{Proximity} = \frac{1}{N} \sum_{i=1}^{N} \mathrm{dist}(\textbf{x}_{i},\textbf{x}_{i}^{\mathrm{cf}})
\end{equation} 
where $\mathrm{dist}(\cdot,\cdot)$ can be any reasonable distance metric such as Euclidean, cosine, or Mahalanobis distance.
An explanation model with lower proximity is preferred, since a smaller distance means that the chosen features of $\mathbf{e}_k^i$ can alter the black-box's output within a more local region of the original input, which means that a slight change on the chosen explanation will change the model prediction, indicating a strong explanation.
In cases where the feature is categorical or the embeddings are not accessible, one can regard this proximity term as being a constant $c$, which is equivalent to assuming $\mathrm{dist}(\mathbf{x}_i,\mathbf{x}_i^{\mathrm{cf}})=c$ for all distances.

Both metrics are commonly used for counterfactual explanation ~\cite{wachter2017counterfactual, molnar2020interpretable}, and as previously described, validity and proximity capture different aspects of the notion.
Thus, we define the overall counterfactual evaluation score (CES) by combining them as Eq.~\eqref{eq:ce}:
\begin{equation}\label{eq:ce}
\begin{aligned}
\small
    \mathcal{C} = \frac{\text{Validity}}{\text{Proximity}} = \frac{\sum_{i=1}^{N} I(\hat{y_{i}} \neq \hat{y_{i}}^{\mathrm{cf}})}{\sum_{i=1}^{N} \mathrm{dist}(\textbf{x}_{i},\textbf{x}_{i}^{\mathrm{cf}})}
\end{aligned}
\end{equation}

With the score calculated as Eq.\eqref{eq:ce} for each explanation method, we say the explainability of $\mathcal{E}_k$ is better than $\mathcal{E}_{k'}$ if $\mathcal{C}(\mathcal{E}_k)>\mathcal{C}(\mathcal{E}_{k'})$ since it indicates either a larger validity score or a smaller proximity score or both.
Additionally, for tasks like binary classification where $\mathcal{F}$ models the label probability, we can provide a finer-grained evaluation result through the softened validity and the softened CES:
\begin{align}
    \small
    \textit{Validity}_\mathrm{soft} & = \frac{1}{N}\sum_{i=1}^{N} \Big(p(\hat{y}_i|\mathbf{x}_i) - p(\hat{y}_i|\mathbf{x}_i^{\mathrm{cf}})\Big)\label{eq:validity_soft}\\
    \small
    \mathcal{C}_{\mathrm{soft}} & = \frac{\sum_{i=1}^{N} \Big(p(\hat{y}_i|\mathbf{x}_i) - p(\hat{y}_i|\mathbf{x}_i^{\mathrm{cf}})\Big)}{\sum_{i=1}^{N} \mathrm{dist}(\mathbf{x}_i,\mathbf{x}_i^{\mathrm{cf}})}\label{eq:ce_soft}
\end{align}
where each term $p(\hat{y}_i|\mathbf{x}_i) -p(\hat{y}_i|\mathbf{x}_i^{\mathrm{cf}})$ represents how the original class probability changes by changing the original input to the counterfactual input.
Here we also point out that Eq.\eqref{eq:validity_soft} and Eq.\eqref{eq:ce_soft} are designed for binary classification task, and may not fully reveal the counterfactual output behavior for multi-class classification. For example, the increase of $p(\hat{y}_i^{\mathrm{cf}}|\mathbf{x}_i^{\mathrm{cf}})$ may not always cause a decrease of the original output $p(\hat{y}_i|\mathbf{x}_i^{\mathrm{cf}})$, they can both be increasing while other classes decrease.


\begin{algorithm}[t]
\textbf{Input:} $\mathbf{x}_i$, $\mathcal{F}$, $\mathcal{E}_k$\\
\textbf{Output:} $\mathbf{x}_i^{\mathrm{cf}}$, $\hat{y}_i^\mathrm{cf}, \mathbf{e}_k^i$ \\
$\hat{y}\leftarrow \mathcal{F}(\mathbf{x}_i)$\\
$\mathbf{e}_k^i\leftarrow$ get explanation by $\mathcal{E}_k$ for $(\mathbf{x}_i,\hat{y}_i)$\\
$S\leftarrow$ search space $V(\mathbf{e}_k^{i,1})\times\dots\times V(\mathbf{e}_k^{i,L_i})$\\
\For{$e'$ in S}{
    $\mathbf{x}_{cf}$ $\leftarrow$ replace corresponding values of $e'$ in $\mathbf{x}_i$\\
    $\hat{y}_i^{cf}$ $\leftarrow$ $\mathcal{F}(\mathbf{x}_{cf})$\\
    \If{$\hat{y}_i^{cf}\neq \hat{y}_i$}{
        Return $\mathbf{x}_i^{\mathrm{cf}}$, $\hat{y}_i^\mathrm{cf}, \mathbf{e}_k^i$ 
    }
}
$e'\leftarrow$ randomly pick one in $\{S \backslash \{e_k^i\}\}$\\
$\mathbf{x}_i^{cf}$ $\leftarrow$ replace values of $e'$ in $\mathbf{x}_i$\\
Return $\mathbf{x}_i$, $\mathcal{F}(\mathbf{x}_i^{cf})$, $\mathbf{e}_k^i$\\
\caption{Discrete Method} 
\label{algo:discrete}
\end{algorithm}

\subsection{Exhaustive Search}\label{sec:discrete}

In order to estimate the metrics in section \ref{sec:ce}, we should first find a counterfactual $\mathbf{x}_i^{\mathrm{cf}}$ that provides the output switch (namely, $\hat{y}_i^{\mathrm{cf}}\neq\hat{y}_i$) for each data point $\mathbf{x}_i$, given the black-box $\mathcal{F}$ and any explanation model $\mathcal{E}_k$.
The most intuitive solution is to consider it as a search problem, and for categorical features, we can apply exhaustive search.

Without loss of generality, assume that all $M$ features of $\mathbf{x}_i$ are categorical, among which $L_i$ ($L_i<M$) features are selected in $\mathbf{e}_k^i$ by $\mathcal{E}_k$.
Denote the search space of the $j$-th feature as $V(\mathbf{e}_k^{i,j})$, then the whole search space would be the Cartesian product $V(\mathbf{e}_k^{i,1})\times\dots\times V(\mathbf{e}_k^{i,L_i})$.
For example, in a human occupation prediction dataset, a record with three categorical features is $\mathbf{x}_i=\{\textit{race}=x_{i,\text{race}}, \textit{gender}=x_{i,\text{gender}},\textit{agegroup}=x_{i,\text{agegroup}}\}$, which has prediction ($\hat{y}_i=\text{student}$) as occupation, then $\mathcal{E}_k$ may return $e_k^i=\{\text{agegroup}=x_{i,\text{agegroup}}\}$ as explanation where $x_{i,\text{agegroup}}=\text{``10-16''}$.
Since there is only one selected feature ($L_i=1$) , finding the counterfactual is simply replacing any different value in $V(\textit{agegroup})$ for the corresponding feature, and see if the modified input results in $\hat{y}_i^{\mathrm{cf}}\neq \text{student}$.
In this example, the discovered counterfactual $\mathbf{x}_i^{\mathrm{cf}}$ may have $\textit{agegroup}=\text{``35-48''}$ while all other features are kept the same.
In cases where there is no counterfactual in the search space that achieves $\hat{y}_i^{\mathrm{cf}}\neq\hat{y}_i$ in the context of $\mathbf{x}_i$, indicating that these features should be less important, we will randomly pick a counterfactual from the search space, and it will have no effect on the final score since $I(\hat{y}_i\neq\hat{y}_i^{\mathrm{cf}})=0$; however, when having access to the predicted probabilities, we will pick the one with the largest $p(\hat{y}_i|\mathbf{x}_i) - p(\hat{y}_i|\mathbf{x}_i^{\mathrm{cf}})$.
The detailed description of this process is given in Algorithm \ref{algo:discrete}.
Note that for the calculation of $\textit{Validity}_{\mathrm{soft}}$ and $\mathcal{C}_{\mathrm{soft}}$, line 15 of the algorithm will return the class probability instead of the predicted label. 
With the returned results from Algorithm \ref{algo:discrete}, we can compute $\mathcal{C}$ or $\mathcal{C}_{\mathrm{soft}}$ by Eq.\eqref{eq:ce} or Eq.\eqref{eq:ce_soft}, and denote them as $\mathcal{C}$(disc.) or $\mathcal{C}_{\mathrm{soft}}$(disc.).


\subsection{Continuous Relaxation}\label{sec:continuous}

Exhaustive search is useful, however, for continuous features or features with a large number of possible values, it may become inefficient.
Thus, we provide a relaxation that considers the search of counterfactuals as an optimization problem.
Denote $w_{i,j} \in R^d$ as the embedding for the $j$-th selected feature in $\mathbf{e}_k^i$, and the dimension size $d$ may vary according to the feature format.
In the case of text classification where $\mathbf{x}_i$ is a word sequence, $w_{i,j}$ corresponds to the word embedding.
Then the optimization problem is formulated as:
\begin{equation} \label{eq:obj}
\small
\begin{aligned}
    \min \sum_{j=1}^{L_i} \|\mathbf{w}_{i,j} - \mathbf{w}_{i,j}^{\mathrm{cf}}\|^2_2 \\
    s.t. \quad \hat{y_i} \neq \mathcal{F}(\mathbf{x}_i^{cf})
\end{aligned}
\end{equation}
where $\mathbf{w}_{i,j}$ and $\mathbf{w}_{i,j}^{\mathrm{cf}}$ are the embeddings of the original and the counterfactual features respectively.
This objective nicely aligns with the evaluation metrics since the $\ell_2$-norm corresponds to the proximity and the constraint corresponds to the validity.

To solve this non-differentiable objective, we further consider that the prediction of the black-box model represents the label probability like that in classification tasks.
Then, we have a relaxed objective function, shown as below,
\begin{equation}\label{eq:continuous}
\small
    \min \sum_{j=1}^{L_i} \|\mathbf{w}_{i,j} - \mathbf{w}_{i,j}^{\mathrm{cf}}\|^2_2 + \alpha \cdot p(\hat{y}_i |\mathbf{x}_i^{cf})
\end{equation}
where $\alpha$ is a trade-off coefficient between the two terms.
While the first term comes from the original objective, the second term in Eq.\eqref{eq:continuous} forces the probability of inferring the same class as the original input to be smaller, which gives higher chance to observe $\hat{y}_i^{\mathrm{cf}}\neq\hat{y}_i$.
Note that $\mathbf{w}_{i,j}^{\mathrm{cf}}$ for selected features in $\mathbf{e}_k^{i}$ are randomly initialized and are the only variables that need optimization.
The total number of parameters needing to be optimized is $Nd$, where $N$ is the number of data points, and $d$ is the embedding size.
If the gradient information $\partial{p}/\partial{\mathbf{w}_{i}}$ is available, one can simply apply gradient-based optimization until it converges.
The whole process can be summarized as Algorithm \ref{algo:continuous}.
With the returned results from Algorithm \ref{algo:continuous}, we can compute $\mathcal{C}$ or $\mathcal{C}_{\mathrm{soft}}$ by Eq.\eqref{eq:ce} or Eq.\eqref{eq:ce_soft}, and denote them as $\mathcal{C}$(cont.) or $\mathcal{C}_{\mathrm{soft}}$(cont.).


\begin{algorithm}[t]
\textbf{Input:} $\mathbf{x}_i, \mathcal{F}, \mathcal{E}_k$\\
\textbf{Output:} $\mathbf{w}_{i}^{\mathrm{cf}}$, $\mathbf{x}_i^{\mathrm{cf}}$, $\hat{y}_i^\mathrm{cf}, \mathbf{e}_k^i$ \\
$\hat{y}\leftarrow \mathcal{F}(\mathbf{x}_i)$\\
$\mathbf{e}_k^i\leftarrow$ get explanation by $\mathcal{E}_k$ for $(\mathbf{x}_i,\hat{y}_i)$\\
$\mathbf{w}_{i}^{\mathrm{cf}}\leftarrow$ solve Eq.~\eqref{eq:continuous}\\
$\mathbf{x}_i^{\mathrm{cf}}\leftarrow$ replace values of $\mathbf{w}_{i}$ in $\mathbf{x}_i$\\
Return $\mathbf{w}_{i}^{\mathrm{cf}}$, $\mathbf{x}_i^{\mathrm{cf}}$, $\mathcal{F}(\mathbf{x}_i^\mathrm{cf})$, $\mathbf{e}_k^i$\\
\caption{Continuous Method} \label{algo:continuous}
\end{algorithm}


\section{Experiments}\label{sec:experiment}

\subsection{Experiment Setup}
We perform experiments for binary classification tasks with text and tabular data\footnote{Code will be released once published, we also provide details about the hyperparameter setting in Appendix \ref{sec:reproducibility}.}. 
The first dataset consists of movie review excerpts ~\cite{10.3115/1118693.1118704}. 
This dataset, \textit{Movie Reviews}, includes 10,443 reviews with binary sentiment labels, which indicate positive or negative tones (among them, positive : negative = 5221 : 5212).
The second dataset is the tabular data \textit{Adults} from the UCI ML repository\footnote{\url{http://archive.ics.uci.edu/ml/datasets/Adult}} ~\cite{Dua:2019}. 
This dataset contains records of 15,682 individuals, and the binary label is whether their annual income is more than \$50,000 (among them, True : False = 7841 : 7841). 
For both of them, we use the same data pre-processing scheme as~\cite{hase2020evaluating,ribeiro2018anchors}, then split into partitions of 80\%, 10\%, and 10\% for the train, validation, and test sets, respectively. 
We use train and validation sets to train and validate the classifier, and run explanation methods to generate explanations on the test set.

\subsection{Whitebox Model}
Similar to \citet{ribeiro2016should}, we measure faithfulness of explanations on a classifier that is by itself interpretable, e.g., logistic regression (LR), and use its results as the ground truth.
In particular, we train two classifiers for both tabular and text classification tasks, and thus we know the gold-standard set of features that are considered important by these models.
The detailed process of how to identify important features of LR and then use them as explanations can be found in~\citet{molnar2020interpretable} (section 4.2\footnote{\url{https://christophm.github.io/interpretable-ml-book/logistic.html}}).
For each prediction in the test set, we use explanation methods to generate explanations and compute the fraction of these gold-standard features that are recovered by these acquired explanations.
We mark fractions as ``GroundTruth*'' in Table \ref{tab:rank} and \ref{tab:multi}, and use them to evaluate the performances of CE and erasure-based metrics.


For tabular data, since each data contains twelve user features (e.g., sex, race, nation, and education, etc.), we simply pass them through an embedding layer, concatenate their embeddings together as the input, and pass it through the LR model.
For text data, we first mask each sentence to a fixed length (e.g., the largest length of sentences in the dataset), pass them through an embedding layer to map each word in GloVe word embedding space~\cite{pennington2014glove}, concatenate all of them, and pass the latent embedding through LR.
Finally, the model test accuracies on two datasets are  77.57\% and 71.20\% for \textit{Adults} and \textit{Movie Reviews}.


\subsection{Explanation Models}
In this section, we describe the explanation methods that are used in our experiment.\\
\textbf{LR}: it is inherently interpretable.\\
\textbf{Random}: A random procedure that is used to randomly pick $K$ features/words as explanations. \\
\textbf{Omission}: It aims to estimate the contribution of a single feature/token by deleting them and evaluating the effect, e.g., by the difference in probability~\cite{4407709}.\\
\textbf{LIME}: \citet{ribeiro2016should} proposed LIME as a local linear approximation of model behavior. 
Using a user-specified feature space, the linear model fits the black-box output on the samples that are distributed around the input. \\
\textbf{Anchor}: 
\citet{ribeiro2018anchors} introduced Anchor, an extension of LIME learning a list of rules that can predict model behavior with high confidence.
An anchor explanation is a decision rule that sufficiently ties a prediction locally that changes to the rest of the features values have no impact.\\
\textbf{Decision Boundary (DB)}: 
\citet{jacovi2020towards} provided a simple method to generate counterfactual explanations. 
They first sampled around the original input to get instances that cross the decision boundary. 
Then, a counterfactual input is chosen from these by taking the instance with the fewest edited features, while breaking ties using the Euclidean distance between latent representations. 

\subsection{Baselines}
As mentioned in related work, there are several erasure-base metrics for faithfulness evaluation.
Among them, we choose the most representative ones---\citet{serrano-smith-2019-attention}
and \citet{deyoung2019eraser}---as our baselines.

First of all, \citet{deyoung2019eraser} defined two simple metrics---comprehensiveness score (Comp.) and sufficiency score (Suff.)---for faithfulness evaluation, which are shown as below.
\begin{align}
\small
\text{Comp.} & = \frac{1}{N} \sum_{i=1}^{N} \left(p(\hat{y}_i|\mathbf{x}_i)-p(\hat{y}_i|\mathbf{x}_i \backslash \mathbf{e}_i)\right)\\
\small
\text{Suff.} & = \frac{1}{N} \sum_{i=1}^{N} \left(p(\hat{y}_i|\mathbf{x}_i)-p(\hat{y}_i|\mathbf{e}_i)\right)
\end{align}
where $\mathbf{x}_i \backslash \mathbf{e}_i$ is the remaining features/tokens in $\mathbf{x}_i$ after deleting those in $\mathbf{e}_i$.

While it would be easy for sequential model to delete features/tokens in the input directly, it may cause problems for LR as the size of the input is fixed.
Therefore, following \citet{zaidan2007using}, we propose two ways to do ``removal''. 
First, replacing it with a zero embedding so that we can block all the effects from that feature.
Second, masking it, for example, replacing it with the embedding of ``<unk>'' from the embedding space, as simply deleting a rationale substring might cause deep feature extraction to behave strangely~\cite{zaidan2007using}.
Finally, in the explanation results, we present both Comp.(del.), Suff.(del.) and Comp.(mask), Suff.(mask) score pairs to represent these two ways, respectively.

Secondly, we adapt the setting in~\citet{serrano-smith-2019-attention} to fit our experimental setting, specifically, we fix the number of features/tokens in each explanation, then count the fraction of decision flip caused by removing the set of explanation.
The definition of \textit{Decision Flip Ratio} (DFR) is:
\begin{equation}
\small
   \text{DFR} = \frac{\sum_{i=1}^{N} I(\hat{y}_i \neq \mathcal{F}(\mathbf{x}_i \backslash \mathbf{e}_i))}{N}
\end{equation}

The higher the fraction of decision flip caused by erasing explanation, the more faithful the explanation provided by an explainable method is.



\begin{table*}
\vspace{-20pt}
\caption{Summary of the evaluation results on two datasets. The ground-truth ranking is marked with a ``*'', and the ranking results consistent with it are marked in bold. Euclidean distance is used to calculate all \textit{Proximity}.}
\centering
\begin{adjustbox}{max width=\linewidth}
\setlength{\tabcolsep}{7pt}
\begin{tabular}
    {m{2.5cm} c c c c c c c c} \toprule
    \multirow{2}{*}{Metric} 
    & \multicolumn{1}{c}{Random } 
    & \multicolumn{1}{c}{Omission } 
    & \multicolumn{1}{c}{LIME } 
    & \multicolumn{1}{c}{Anchor } 
    & \multicolumn{1}{c}{DecisionBoundary} 
    & \multicolumn{1}{c}{LogisticRegression} 
    \\\cmidrule(lr){2-2} \cmidrule(lr){3-3} \cmidrule(lr){4-4} \cmidrule(lr){5-5} \cmidrule(lr){6-6} \cmidrule(lr){7-7}
    \multicolumn{7}{c}{Adults} \\\midrule
    Comp. (del.) $\uparrow$ & 0.0217 (\textbf{6th}) & 0.0314 (\textbf{5th}) & 0.1228 (4th) & 0.1372 (3rd) & 0.1396 (2ed) & 0.1661 (\textbf{1st}) \\
    Suff. (del.) $\downarrow$ & 0.2744 (\textbf{6th}) & 0.2546 (\textbf{5th}) & 01536 (4th) & 0.1483 (3rd) & 0.1466 (2ed) & 0.1096 (\textbf{1st}) \\
    DFR $\uparrow$ & 0.0587 (5th) & 0.0436 (6th) & 0.2378 (4th) & 0.3163 (3rd) & 0.3812 (1st) & 0.3807 (2ed) \\\cmidrule(lr){1-7}
    Validity $\uparrow$ & 0.0682 (\textbf{6th}) & 0.0689 (\textbf{5th}) & 0.2844 (\textbf{2ed}) & 0.2073 (\textbf{4th}) & 0.2736 (\textbf{3rd}) & 0.3565 (\textbf{1st})\\
    Proximity & 1.414 & 1.414 & 1.414 & 1.414 & 1.414 & 1.414 \\
    $\mathcal{C}$ (disc.) $\uparrow$ & 0.0483 (\textbf{6th}) & 0.0487 (\textbf{5th}) & 0.2012 (\textbf{2ed}) & 0.1466 (\textbf{4th}) & 0.1935 (\textbf{3rd}) & 0.2521 (\textbf{1st}) \\\cmidrule(lr){1-7}
    $\text{Validity}_{\mathrm{soft}}$ $\uparrow$ & 0.0196 (\textbf{6th}) & 0.0274 (\textbf{5th}) & 0.1148 (\textbf{2ed}) & 0.1118 (3rd) & 0.1105 (4th) & 0.1721 (\textbf{1st})\\
    Proximity & 2.8455 & 2.9933 & 2.4867 & 3.9865 & 2.4383 & 2.5799 \\
    $\mathcal{C}_{\mathrm{soft}}$ (disc.) $\uparrow$ & 0.0069 (\textbf{6th}) & 0.0092 (\textbf{5th}) & 0.0462 (\textbf{2ed}) & 0.0281 (\textbf{4th})& 0.0453 (\textbf{3rd}) & 0.0667 (\textbf{1st})\\\cmidrule(lr){1-7}
    GroundTruth* $\uparrow$ & 0.0746 (6th*) & 0.1033 (5th*) & 0.6466 (2ed*) & 0.2334 (4th*) & 0.3173 (3rd*) & 1.0 (1st*)\\\midrule

\multicolumn{7}{c}{Movie Reviews} \\\midrule
    Comp. (del.) $\uparrow$ & 0.0376 (6th) & 0.0483 (4th) & 0.0868 (\textbf{3rd}) & 0.1956 (\textbf{2ed}) & 0.0391 (5th) & 0.2272 (\textbf{1st})\\
    Suff. (del.) $\downarrow$ & 0.4325 (4th) & 0.4377 (5th) & 0.4233 (\textbf{3rd}) & 0.4140 (\textbf{2ed}) & 0.4337 (6th) & 0.4121 (\textbf{1st})\\
    Comp. (mask) $\uparrow$ & 0.0358 (6th) & 0.0499 (4th) & 0.0872 (\textbf{3rd}) & 0.2162 (\textbf{2ed}) & 0.0432 (5th) & 0.2225 (\textbf{1st})\\
    Suff. (mask) $\downarrow$ & 0.4325 (4th) & 0.4377 (5th) & 0.4233 (\textbf{3rd}) & 0.4140 (\textbf{2ed}) & 0.4337 (6th) & 0.4121 (\textbf{1st})\\
    DFR $\uparrow$ & 0.0838 (\textbf{5th}) & 0.0972 (4th) & 0.1365 (\textbf{3rd}) & 0.2566 (\textbf{2ed}) & 0.0765 (6th) & 0.2898 (\textbf{1st})\\\cmidrule(lr){1-7}
    Validity $\uparrow$ & 0.8743 (6th) & 0.8920 (5th) & 0.9102 (\textbf{3rd}) & 0.9197 (\textbf{2ed}) & 0.8944 (\textbf{4th})& 0.1099 (\textbf{1st})\\
    Proximity & 0.1573 & 0.1405 & 0.1194 & 0.1088 & 0.1376 & 0.0775 \\
    $\mathcal{C}$ (cont.) $\uparrow$ & 5.5590 (6th) & 6.3479 (5th) & 7.6240 (\textbf{3rd}) & 8.4492 (\textbf{2ed}) & 6.5007 (\textbf{4th})& 12.2585 (\textbf{1st})\\\cmidrule(lr){1-7}
    $\text{Validity}_{\mathrm{soft}}$ $\uparrow$ & 0.7769 (6th) & 0.7953 (5th) & 0.8147 (\textbf{3rd}) & 0.8216 (\textbf{2ed}) & 0.7969 (\textbf{4th}) & 0.8517 (\textbf{1st})\\
    Proximity & 0.1573 & 0.1405 & 0.1194 & 0.1088 & 0.1376 & 0.0775 \\
    $\mathcal{C}_{\mathrm{soft}}$ (cont.) $\uparrow$ & 6.3586 (6th) & 7.1169 (5th) & 8.3763 (\textbf{3rd}) & 9.1873 (\textbf{2ed}) & 7.2681 (\textbf{4th}) & 12.9077 (\textbf{1st})\\\cmidrule(lr){1-7}
    GroundTruth* $\uparrow$ & 0.0790 (5th*) & 0.0492 (6th*) & 0.3065 (3rd*) & 0.539 (2ed*)& 0.0946 (4th*) & 1.0 (1st*) \\\bottomrule
\end{tabular}
\end{adjustbox}
\label{tab:rank}
\vspace{-10pt}
\end{table*}

\begin{table}
\centering
\caption{Summary of correlations between each evaluation result and the ground-truth result on two datasets. Each value in this table is computed using Eq.\eqref{eq:tau} or Eq.\eqref{eq:rho}. The largest values in each column are in bold.}
\begin{adjustbox}{max width=\linewidth}
\setlength{\tabcolsep}{7pt}
\begin{tabular}
    {m{2.5cm} c c c c} \toprule
    \multirow{2}{*}{Metric} 
    & \multicolumn{2}{c}{Adults} 
    & \multicolumn{2}{c}{Movie Reviews}\\
    \cmidrule(lr){2-3} \cmidrule(lr){4-5}
    & \multicolumn{1}{c}{$\tau$ $\uparrow$} 
    & \multicolumn{1}{c}{$\rho$ $\uparrow$} 
    & \multicolumn{1}{c}{$\tau$ $\uparrow$} 
    & \multicolumn{1}{c}{$\rho$ $\uparrow$}\\ \midrule
    Comp. (mask)  & 0 & 0 & 0.7333 & 0.8285  \\
    Suff. (mask)  & 0 & 0 & 0.6 & 0.7714 \\
    Comp. (del.) & 0.7333 & 0.8285 & 0.7333 & 0.8285\\
    Suff. (del.)  & 0.7333 & 0.8285 & 0.6 & 0.7714\\
    DFR  & 0.4667 & 0.6571 & 0.6 & 0.7714\\
    $\mathcal{C}$ (disc.) & \textbf{10} & \textbf{1.0} & 0 & 0\\
    $\mathcal{C}$ (cont.) & 0 & 0 & \textbf{0.8666} & \textbf{0.9428}\\
    $\mathcal{C}_{\mathrm{soft}}$ (cont.) & \textbf{1.0} & \textbf{1.0} & \textbf{0.8666} & \textbf{0.9428}\\
\bottomrule
\end{tabular}
\end{adjustbox}
\label{tab:result}
\vspace{-16pt}
\end{table}

\begin{table}[h]
\caption{Summary of correlations between each evaluation result and the ground-truth result in \textit{Adults} given two explanations (features/words). For erasure-based metrics, we remove both features directly; for CE, we find counterfactuals in the combinations of them.}
\begin{adjustbox}{max width=\linewidth}
\centering
\begin{tabular}
    {m{1cm} c c c c c} \toprule
    \multirow{2}{*}{Metric} 
    & \multicolumn{1}{c}{Comp. (del.) $\uparrow$} 
    & \multicolumn{1}{c}{Suff. (del.) $\downarrow$} 
    & \multicolumn{1}{c}{DFR. $\uparrow$} 
    & \multicolumn{1}{c}{$\mathcal{C}_{soft}$ (cont.)$\uparrow$} 
    & \multicolumn{1}{c}{GroundTruth* $\uparrow$} 
    \\\cmidrule(lr){2-3} \cmidrule(lr){4-4} \cmidrule(lr){5-5} \cmidrule(lr){6-6}
    \multicolumn{6}{c}{Adults} \\\midrule
   
Random   &  0.0440 (\textbf{4th}) & 0.2476 (\textbf{4th}) & 0.1116 (\textbf{4th}) & 0.0148 (\textbf{4th})  & 0.2991 (4th*)\\
LIME   &  0.1751 (\textbf{3rd}) & 0.1024 (\textbf{3rd}) & 0.3807 (\textbf{3rd}) & 0.0699 (\textbf{3rd})  & 0.5433 (3rd*)\\
Anchor  &  0.2876 (1st) & -0.0141 (1st) & 0.5332 (1st) & 0.0702 (\textbf{2ed})  & 0.7429 (2ed*)\\
LR  &  0.2648 (2ed) & 0.0066 (2ed)& 0.5198 (2ed) & 0.0775 (\textbf{1st})  & 1.0 (1st*)\\\midrule
$\tau$ $\uparrow$ &  0.6667  & 0.6667 & 0.6667  & \textbf{1.0}  & \\
$\rho$ $\uparrow$ &  0.8000  & 0.8000  & 0.8000  & \textbf{1.0}  & \\
\bottomrule
\end{tabular}
\end{adjustbox}
\label{tab:multi}
\end{table}

\subsection{Measurements}
We use two non-parametric methods for measuring the correlation between evaluation methods and ground-truth: Kendall’s $\tau$ and Spearman’s $\rho$.\\
\textbf{Kendall's $\tau$}~\cite{10.1093/biomet/33.3.239} is a statistical metric used to measure the ordinal association between two measured quantities. 
Considering the similarity of orderings of $u_1, \dots, u_n$ and $v_1, \dots, v_n$. 
For any pair of indices $1 \leq i < j \leq n$: if both $u_i > u_j$ and $v_i > v_j$, or both $u_i < u_j$ and $v_i < v_j$, $(i, j)$ is called a concordant pair; if both $u_i > u_j$ and $v_i < v_j$, or both $u_i < u_j$ and $v_i > v_j$, $(i, j)$ is called a discordant pair; if $u_i = u_j$ or $v_i = v_j$, $(i, j)$ is neither concordant nor discordant.
Then, Kendall’s $\tau$ is defined as,
    \begin{equation}\label{eq:tau}
    \small
    \tau_{u v}=\frac{|\text{concordant pairs}|-|\text {discordant pairs}|}{n(n-1) / 2}
    \end{equation}
    \\
\textbf{Spearman's $\rho$} ~\cite{zwillinger1999crc} is a non-parametric measure of rank correlation (statistical dependence between the rankings of two variables). 
It assesses how well the relationship between two variables can be described using a monotonic function.
To compute $\rho$, we first compute the ranks of the sample values, then replace $u_1$, \dots, $u_n$ with ranks $a_1$, \dots, $a_n$, 
and replace $v_1$, \dots, $v_n$ with ranks $b_1$, \dots, $b_n$.
We assign smaller ranks to smaller $u$ and $v$. 
Then, Spearman's $\rho$ is
    \begin{equation}\label{eq:rho}
    \small
    \rho_{u v}=1-\frac{6 \sum_{i=1}^{n}\left(a_{i}-b_{i}\right)^{2}}{n\left(n^{2}-1\right)}
    \end{equation}

\subsection{Results}

The major experimental results are shown in Table \ref{tab:rank}, Table \ref{tab:result} and Table \ref{tab:multi}, while Table \ref{tab:rank} presents the evaluation value for each explanation method based on different measurements, Table \ref{tab:result} shows the results which are calculated using Eq.\eqref{eq:tau} and Eq.\eqref{eq:rho} based on the ranking results from Table \ref{tab:rank}, and Table \ref{tab:multi} is the results when measuring multiple explanations.
For $\textit{Adults}$, we only present Comp.(del.) and Suff.(del.), as the features are categorical, which is not suitable for masking; for \textit{Movie Reviews}, we only present $\mathcal{C}$(cont.) and $\mathcal{C}_{\mathrm{soft}}$(cont.) given the large search space.
We analyze and discuss the results in terms of the following perspectives.

\subsubsection*{\bf 1) Only Having Predicted Labels}
If we can only get the labels predicted by the model, then only CE using brute search ($\mathcal{C}$ (disc.)) and DFR~\cite{serrano-smith-2019-attention} can be used to measure faithfulness, as the computation of comprehensiveness and sufficiency needs predicted probabilities for each label.
From Table \ref{tab:result}, we can see that the $\mathcal{C}$(disc.) has much higher values of $\tau$
and $\rho$ than DFR, indicating a much stronger correlation with the ground-truth results.
When checking the detailed values in Table \ref{tab:rank}, we can easily find that the ranking result provided by DFR for \textit{Adults} is completely inconsistent with the ground-truth ranking, while $\mathcal{C}$ (disc.) is 100\% correct.
Besides, the reason that \textit{Proximity} of $\mathcal{C}$(disc.) are all 1.41 is we use one-hot embedding to represent the categorical features, and use Euclidean distance to compute.


\subsubsection*{\bf 2) Having Probabilities \& Embeddings}
Once we have access to the predicted probability for each label from the black-box model, $\mathcal{C}$ (cont.), $\mathcal{C}_{\mathrm{soft}}$(cont.), comprehensiveness (Comp.) and sufficiency (Suff.) can be used to measure.
From Table \ref{tab:result}, we can find that Comp.(del.) is the strongest baseline, but $\mathcal{C}$ (cont.) still achieves much higher values of $\tau$ and $\rho$ compared with it on both datasets, and achieves an average improvement of 17.83\%.
Besides, as is shown in Table \ref{tab:result}'s $\textit{Movie Reviews}$, Comp. and Suff. do not always return the same results, once this happened, it would become even harder for users to decide which one to trust.
However, our proposed method CE will not get into this predicament.
One more interesting observation is that,
for \textit{Adults} dataset and  $\mathcal{C}_{\mathrm{soft}}$ in Table \ref{tab:rank}, we can see that the ranking result has been significantly refined with the help from \textit{Proximity} compared with the result of $\textit{Validity}_{\mathrm{soft}}$, which, to some extent, shows the rationality of our choice of \textit{Proximity} in the definition of $\mathcal{C}$ and $\mathcal{C}_{\mathrm{soft}}$.

\subsubsection*{\bf 3) Having Multiple Explanations}
Table \ref{tab:multi} shows the results when evaluation two explanations.
We only choose Random, LIME, and Anchor in $\textit{Adults}$ dataset, as other algorithms can hardly provide more than one explanations on both datasets (less than 50\%) and Anchor can hardly provide more than one explanations on \textit{Movie Reviews} (less than 50\%).
As is shown in Table \ref{tab:multi}, the performance of $\mathcal{C}_{\mathrm{soft}}$ is still far beyond the other two baselines, and is 100\% consistent with the ground-truth ranking result.


\section{Conclusion}
In this paper, we propose a new methodology to evaluate the faithfulness of explanations through the lens of causality, known as counterfactual evaluation.
CE can be regarded as a generalization of the existing erasure-based criteria, and at the same time, it mitigates the intrinsic bias of erasure-based criteria by introducing counterfactual samples.
Empirical experiments also verify the superiority of CE as it achieves the highest consistency with ground-truth compared with other metrics.

\newpage
\bibliography{anthology,custom}
\bibliographystyle{acl_natbib}

\end{document}